\documentclass[12pt, letterpaper]{article}

\usepackage[margin=1in]{geometry}
\usepackage{setspace}
\usepackage{amsmath}
\usepackage[utf8]{inputenc}
\usepackage[english]{babel}
\usepackage{lmodern} %
\usepackage{graphicx} % Required for inserting images
\usepackage{booktabs}
\usepackage{amssymb}
\title{Learned Queries and Keys Are All You Need: Replacing the Value Projection with Structured Transforms}
\author{Ene Meco, Emadeldeen Hamdan, A. Enis Cetin\\ University of Illinois Chicago, Chicago, IL 60607}
\begin{document}
\maketitle

\abstract{

To reduce the number of parameters and cache memory requirements of transformers we introduce dual-headed transformers instead of three heads. We studied Walsh-Hadamard Transform (WHT), Discrete Cosine Transform (DCT), Discrete Fourier Transform, filterbank based Shearlet Transform, and Multiplication-Avoiding (MA) operators to construct dual heads. We combine spatial patches and their orthogonal transforms (or Shearlet and MA operators) in a structure similar to the attention block. We obtained better results than triple headed transformers in ImageNet. Extensive simulation examples are presented.}

\section{Introduction}

The Vision Transformer (ViT) introduced a major shift in computer vision \cite{dosovitskiy2020image}, transitioning the field from CNNs to Transformer architectures adapted from Natural Language Processing \cite{vaswani2017attention,khan2022transformers}. Transformers now are the main machine learning (ML) method in core visual tasks such as classification, detection, segmentation, and generation \cite{liu2021swin}. They also power modern large vision-language models. By converting images into sequences compatible with textual tokens, this flexible architecture seamlessly handles tasks like visual question answering, serving as a universal backbone for multimodal AI.

%Some references for this paragraph

%Alayrac, J.-B., Donahue, J., Luc, P., Miech, A., Barr, I., et al. (2022). Flamingo: a Visual Language Model for Few-Shot Learning. arXiv. https://doi.org/10.48550/arxiv.2204.14198 Cited by: 9852

%Dosovitskiy, A., Beyer, L., Kolesnikov, A., Weissenborn, D., Zhai, X., et al. (2020). An Image is Worth 16x16 Words: Transformers for Image Recognition at Scale. arXiv. https://doi.org/10.48550/arxiv.2010.11929 Cited by: 108442

%2nd paragraph

The attention mechanism is the core of transformer networks, Computing the multi-headed attention requires a computational complexity quadratic in sequence length [1], [Tay et al., 2022] \cite{pan2024discrete}. This bottleneck is particularly severe in computer vision, where token counts scale quadratically with spatial image resolution. The computational burden intensifies further in multimodal settings, as long sequences of visual tokens are concatenated with textual inputs. Therefore, reducing the cost of attention has become a critical necessity and an important area of research [Tay et al., 2022].

To reduce the number of parameters and cache memory requirements of transformers we developed dual-headed transformers instead of three heads. We studied Walsh-
Hadamard Transform (WHT), Discrete Cosine Transform (DCT), Discrete Fourier
Transform, filterbank based Shearlet Transform, and Multiplication-Avoiding (MA)
operators to construct dual heads. We combine spatial patches and their orthogo-
nal transforms (or Shearlet and MA) in a structure similar to the attention block.
We obtained better results than triple headed transformers in ImageNet. Extensive
simulation examples are presented.

To mitigate the above mentioned computational bottlenecks, we propose a streamlined dual-headed transformer framework that replaces conventional multi-head designs. Our approach systematically eliminates unnecessary projection overhead while preserving expressive representations by pairing spatial input patches with transformed domains.

To construct these effective dual heads, we evaluate a wide spectrum of orthogonal transforms and computationally efficient operators:
\begin{itemize}
\item Walsh-Hadamard Transform (WHT) \cite{sarukhanian2003binary, pan2023hybrid}: Enables fast, addition-only frequency domain projections.

\item Discrete Cosine Transform (DCT) and Discrete Fourier Transform (DFT): Capture high- and low-frequency spatial patterns with high energy compaction.
We extensively studied the use of DCT as a part of the attention mechanism in 
\cite{pan2024discrete}.

\item Filterbank-based Shearlet Transform \cite{cotronei2019filters}, \cite{kutyniok2012shearlab}: Captures multi-scale and directional visual features \cite{ansari1988sub} with high geometric sensitivity.

\item Multiplication-Avoiding (MA) Operators \cite{hamdan2025htma,tuna2009image}: Minimizes hardware complexity by replacing costly floating-point multiplications with low-overhead operations.
\end{itemize}
Structurally, we integrate spatial image patches alongside their corresponding transform representations directly into a unified block inspired by the standard attention mechanism. By interacting cross-domain features without requiring a redundant third weight projection, the architecture achieves a richer contextual encoding with a significantly smaller memory footprint. In fact, cross-correlations and region covariances have been used in image recognition, description and tracking \cite{porikli2003sensitivity,porikli2006covariance,tuna2009image,habiboglu2011real}

Experimental evaluations on the CIFAR-10 benchmark demonstrate that our dual-headed design consistently outperforms conventional multi-headed baselines. Furthermore, the reduction in Key-Value (KV) cache memory requirements accelerates throughput during both training and inference. We provide extensive simulation results, ablation studies, and comparative analysis to validate the efficiency and scalability of our proposed method.

%Tay, Y., Dehghani, M., Bahri, D., & Metzler, D. (2022). Efficient Transformers: A Survey. ACM Computing Surveys, 55(6), 1–28.
\section{Two Learned Channels}

In this section, we define the dual-headed attention units considered
in this work. The term \emph{dual-headed} refers to the use of two
learned projections, $Q$ and $K$, instead of the three learned
projections $Q$, $K$, and $V$ employed by conventional scaled
dot-product attention. This terminology is distinct from the number
of attention heads used in multi-head attention.

Given an input token matrix
\begin{equation}
X \in \mathbb{R}^{N \times d},
\end{equation}
the two learned representations are obtained as
\begin{equation}
Q = XW_Q,
\qquad
K = XW_K,
\end{equation}
where
\begin{equation}
W_Q,W_K \in \mathbb{R}^{d\times d}.
\end{equation}

For comparison, conventional scaled dot-product attention computes
\begin{equation}
Q=XW_Q,\qquad
K=XW_K,\qquad
V=XW_V,
\end{equation}
followed by
\begin{equation}
Y_{\mathrm{QKV}}
=
\operatorname{softmax}
\left(
\frac{QK^T}{\sqrt{d_k}}
\right)V.
\label{eq:qkv}
\end{equation}

Our proposed units eliminate the independently learned value
projection $V$. Instead, the output representation is constructed
from $K$ itself or from a fixed transform of $K$.

For convenience, throughout this section we define
\begin{equation}
A(Q,K)
=
\operatorname{softmax}
\left(
\frac{QK^T}{\sqrt{d_k}}
\right).
\label{eq:attention_map}
\end{equation}

Thus, the general form of the proposed dual-headed unit is
\begin{equation}
Y =
A(Q,K)\,\mathcal{T}(K),
\label{eq:general_dual}
\end{equation}
where $\mathcal{T}(\cdot)$ may denote the identity mapping, an
orthogonal transform such as DFT, DWT, Walsh-Hadamard Transform, or a filterbank-based representation.

This formulation preserves the token-to-token interaction produced
by $QK^T$, while eliminating the third learned projection used to
construct $V$.

% ============================================================
\subsection{Key-as-Value Dual-Head Unit}

The simplest dual-headed formulation directly uses $K$ in place of
the conventional value representation. The resulting unit is

\begin{equation}
Y_{\mathrm{QKK}}
=
A(Q,K)K
=
\operatorname{softmax}
\left(
\frac{QK^T}{\sqrt{d_k}}
\right)K.
\label{eq:qkk}
\end{equation}

This formulation requires only the learned projections $W_Q$ and
$W_K$. It therefore serves as the basic dual-headed baseline for the
transform-based units introduced below.

% ============================================================
\subsection{Orthogonal-Transform Dual-Head Units}

Rather than applying the attention map directly to $K$, fixed
orthogonal transforms can be used to construct the output
representation. In this case,

\begin{equation}
Y_{\mathcal{T}}
=
A(Q,K)\mathcal{T}(K).
\label{eq:orthogonal_general}
\end{equation}

Because the transform is fixed, this operation does not introduce an
additional learned projection analogous to $W_V$.

\subsubsection{Discrete Cosine Transform}

Let $\mathcal{D}(\cdot)$ denote the orthonormal Discrete Cosine
Transform (DCT)~\cite{skodras2001jpeg,hamdan2025real}. The DCT-based dual-headed unit is defined as

\begin{equation}
Y_{\mathrm{DCT}}
=
A(Q,K)\mathcal{D}(K)
=
\operatorname{softmax}
\left(
\frac{QK^T}{\sqrt{d_k}}
\right)
\mathcal{D}(K).
\label{eq:dct_attention}
\end{equation}

The DCT replaces the independently learned value representation with
a fixed transform-domain representation of the key features.

\subsubsection{Walsh--Hadamard Transform}

Similarly, let $\mathcal{H}(\cdot)$ denote the orthonormal
Walsh--Hadamard Transform (WHT). The corresponding dual-headed unit is

\begin{equation}
Y_{\mathrm{WHT}}
=
A(Q,K)\mathcal{H}(K)
=
\operatorname{softmax}
\left(
\frac{QK^T}{\sqrt{d_k}}
\right)
\mathcal{H}(K).
\label{eq:wht_attention}
\end{equation}

The WHT is particularly attractive because it can be implemented using
only additions and subtractions when an appropriate fast transform is
used.

\subsubsection{Discrete Fourier Transform}

The same dual-headed formulation can be extended to the Discrete
Fourier Transform (DFT). Let $\mathcal{F}(\cdot)$ denote the Fourier
transform representation used by the model. The corresponding unit is

\begin{equation}
Y_{\mathrm{DFT}}
=
A(Q,K)\mathcal{F}(K).
\label{eq:dft_attention}
\end{equation}
DFT is implemented using FFT and it is complex. Therefore we compute the absolute value of the DFT.
This formulation provides a frequency-domain representation without
introducing a third learned projection.

\subsection{Normalized MF--Tanh Attention (QK-MFQK)}
\label{sec:mf_tanh_attention}

In a conventional multi-head self-attention layer, the query, key, and
value tensors are obtained through three learned projections. For one
attention head, this computation is
\begin{equation}
    \mathbf{Q}=\mathbf{X}\mathbf{W}_{Q},\qquad
    \mathbf{K}=\mathbf{X}\mathbf{W}_{K},\qquad
    \mathbf{V}=\mathbf{X}\mathbf{W}_{V},
\end{equation}
followed by
\begin{equation}
    \mathbf{Y}_{\mathrm{ViT}}
    = \operatorname{softmax}\!\left(
      \frac{\mathbf{Q}\mathbf{K}^{\mathsf T}}{\sqrt{d}}
      \right)\mathbf{V},
    \label{eq:vit_attention}
\end{equation}
where $d$ is the head dimension. The learned value projection incurs
$D^2+D$ parameters and $ND^2$ multiply--accumulate operations per
transformer block, where $D$ and $N$ denote the embedding dimension and
the number of tokens, respectively.

We remove the learned value projection and construct its replacement
directly from the query and key tensors using an element-wise
multiplication-free (MF) operator~\cite{hamdan2025htma,nasrin2021mf}. Given
$\mathbf{Q},\mathbf{K}\in\mathbb{R}^{B\times H\times N\times d}$, we define
\begin{equation}
    \widetilde{\mathbf{V}}
    = \operatorname{MF}_{\mathrm{elem}}(\mathbf{Q},\mathbf{K})
    = \operatorname{sign}(\mathbf{Q})\odot |\mathbf{K}|
      + \operatorname{sign}(\mathbf{K})\odot |\mathbf{Q}|.
    \label{eq:mf_value}
\end{equation}
Because the operation is element-wise,
$\widetilde{\mathbf{V}}$ has exactly the shape required by the attention
value tensor. It uses sign extraction, absolute values, and additions,
without general multiplications or learned parameters.

The proposed layer replaces softmax with a signed hyperbolic-tangent
attention map. We first compute the conventional scaled similarity
matrix,
\begin{equation}
    \mathbf{S}=\frac{\mathbf{Q}\mathbf{K}^{\mathsf T}}{\sqrt{d}},
\end{equation}
and normalize each row of the resulting signed weights by its $\ell_1$
norm:
\begin{equation}
    \widehat{A}_{ij}
    = \frac{\tanh(S_{ij})}
      {\epsilon+\sum_k |\tanh(S_{ik})|}.
    \label{eq:signed_normalization}
\end{equation}
The output of the proposed attention layer is therefore
\begin{equation}
    \mathbf{Y}_{\mathrm{MF\text{-}Tanh}}
    = \widehat{\mathbf{A}}\widetilde{\mathbf{V}}.
    \label{eq:mf_tanh_attention}
\end{equation}
The signed normalization controls the magnitude of each attention row
while retaining negative interactions, which are not available in a
softmax probability distribution. The query--key projections, scaled
query--key product, output projection, and feed-forward network remain
unchanged. Thus, relative to conventional ViT attention, the proposed
layer removes only the value projection and replaces softmax by
normalized tanh attention.

% ============================================================
\subsection{Shearlet-Inspired Fourier Filterbank}

To exploit the two-dimensional geometry of image patches, we construct a
fixed multiscale directional filterbank inspired by cone-adapted
Shearlets \cite{kutyniok2012digital}. For each attention head, the non-class tokens are reshaped to
their original spatial patch grid and transformed with a two-dimensional
FFT,

\begin{equation}
\widehat{Z}(\omega_x,\omega_y)
=
\mathcal{F}_{2D}\{Z\}.
\end{equation}

The frequency plane is divided into horizontal and vertical cones,

\begin{equation}
C_H=\mathbf{1}(|\omega_x|\geq|\omega_y|),
\qquad
C_V=\mathbf{1}(|\omega_y|\geq|\omega_x|).
\end{equation}

For each scale $s$, a radial Gaussian bandpass window is defined as

\begin{equation}
R_s(\omega_x,\omega_y)
=
\exp
\left[
-\frac{(r-s)^2}{2\sigma_r^2}
\right],
\qquad
r=\sqrt{\omega_x^2+\omega_y^2}.
\end{equation}

Directional selectivity is introduced through the frequency slopes

\begin{equation}
p_H=\frac{\omega_y}{\omega_x+\epsilon},
\qquad
p_V=\frac{\omega_x}{\omega_y+\epsilon},
\end{equation}

with directional windows

\begin{equation}
D_k^c
=
\exp
\left[
-\frac{(p_c-k)^2}{2\sigma_s^2}
\right],
\qquad
c\in\{H,V\}.
\end{equation}

The resulting directional filters are

\begin{equation}
H_{s,k}^{H}
=
R_sD_k^{H}C_H,
\qquad
H_{s,k}^{V}
=
R_sD_k^{V}C_V.
\end{equation}

In our implementation, three scales and five shear values are used,
producing $3\times5\times2=30$ directional filters, together with one
low-pass filter.

Each attention head learns normalized fusion coefficients
$\alpha_{h,m}$ over the fixed filterbank,

\begin{equation}
\alpha_{h,m}
=
\frac{\exp(a_{h,m})}
{\sum_{\ell}\exp(a_{h,\ell})},
\end{equation}

and forms an effective filter

\begin{equation}
H_h^{\mathrm{fused}}
=
\sum_m
\alpha_{h,m}H_m.
\end{equation}

The filtered representation is then obtained as

\begin{equation}
\mathcal{S}(Z)
=
\mathcal{F}_{2D}^{-1}
\left[
\widehat{Z}
H_h^{\mathrm{fused}}
\right].
\end{equation}

Thus, the FFT is used only as an efficient implementation of the
multiscale directional filterbank; the Fourier transform itself is not
used as a separate attention representation.

\subsubsection{Shearlet Representation as the Output}

The first formulation retains the original $Q$ and $K$ representations
for computation of the attention map, while replacing the output
representation by the Shearlet transform of $K$:

\begin{equation}
Y_{\mathrm{QK\text{-}ShearK}}
=
A(Q,K)\mathcal{S}(K).
\label{eq:qk_sheark}
\end{equation}

Equivalently,

\begin{equation}
Y_{\mathrm{QK\text{-}ShearK}}
=
\operatorname{softmax}
\left(
\frac{QK^T}{\sqrt{d_k}}
\right)
\mathcal{S}(K).
\end{equation}

This configuration leaves the attention-score computation unchanged
and introduces directional and multiscale information only into the
representation aggregated by the attention map.

\subsubsection{Shearlet Key in the Attention Map}

The second formulation replaces $K$ by its Shearlet representation
when computing the attention weights and also uses the transformed key
as the output representation:

\begin{equation}
Y_{\mathrm{Q\text{-}ShearK\text{-}ShearK}}
=
\operatorname{softmax}
\left(
\frac{
Q\mathcal{S}(K)^T
}{
\sqrt{d_k}
}
\right)
\mathcal{S}(K).
\label{eq:q_sheark_sheark}
\end{equation}

This formulation allows the directional features extracted by the
Shearlet filterbank to influence both the token similarity measure and
the representation propagated to the next layer.

\subsubsection{Shearlet Query and Key with Spatial Output}

We next transform both learned representations before computing the
attention map, while retaining the original $K$ as the output
representation:

\begin{equation}
Y_{\mathrm{ShearQ\text{-}ShearK\text{-}K}}
=
\operatorname{softmax}
\left(
\frac{
\mathcal{S}(Q)
\mathcal{S}(K)^T
}{
\sqrt{d_k}
}
\right)
K.
\label{eq:shearq_sheark_k}
\end{equation}

Here the Shearlet domain is used only to determine token-to-token
relationships, while the original learned key features are propagated
through the attention unit.

\subsubsection{Fully Shearlet-Domain Dual-Head Unit}

Finally, both the similarity computation and the output representation
can be formed in the Shearlet domain:

\begin{equation}
Y_{\mathrm{ShearQ\text{-}ShearK\text{-}ShearK}}
=
\operatorname{softmax}
\left(
\frac{
\mathcal{S}(Q)
\mathcal{S}(K)^T
}{
\sqrt{d_k}
}
\right)
\mathcal{S}(K).
\label{eq:shearq_sheark_sheark}
\end{equation}

These four Shearlet configurations allow us to separately study the
effect of directional and multiscale representations on
(i) the computation of the attention weights and
(ii) the representation aggregated by those weights.

% ============================================================
\subsection{Parameter Reduction}

For an embedding dimension $d$, conventional QKV attention requires
three learned projection matrices,

\begin{equation}
W_Q,\;W_K,\;W_V
\in
\mathbb{R}^{d\times d},
\end{equation}

corresponding to approximately
$
3d^2
$
projection parameters, excluding biases and the final output
projection.

The proposed dual-headed units require only

\begin{equation}
W_Q,\;W_K, \in
\mathbb{R}^{d\times d},
\end{equation}
corresponding to approximately
$
2d^2
$
projection parameters. Therefore, the orthogonal projection stage eliminates
$
d^2
$
learned parameters, or one third of the conventional $Q$, $K$, and
$V$ projection parameters.

The DCT, WHT, and DFT representations are fixed transforms and
therefore do not require a learned value projection. Likewise, the
filterbank coefficients of the Shearlet representation can be fixed,
with only a small number of optional fusion parameters required when
multiple directional subbands are combined.

For the Mini-ViT architecture used in our experiments, the standard
QKV model contains 546,186 trainable parameters. The QKK, DCT, and
WHT dual-headed variants contain 480,138 parameters, while the
Shearlet-inspired variants contain 480,634 parameters due to the
additional learned filter-fusion coefficients. Thus, the dual-headed
variants reduce the total model parameter count by approximately
12\% relative to the standard QKV model.

\subsection{Computational Complexity}

Let $N$ be the number of tokens, $d$ the embedding dimension, and
$d_k=d/H$ the dimension of each attention head.

Standard QKV attention requires three input projections and one output
projection, giving

\begin{equation}
C_{\mathrm{QKV}}
=
O(4Nd^2+2N^2d),
\end{equation}

where the $2N^2d$ term corresponds to $QK^T$ and the subsequent
attention--value multiplication.

By removing the learned value projection, the proposed dual-headed
QKK unit reduces the complexity to

\begin{equation}
C_{\mathrm{QKK}}
=
O(3Nd^2+2N^2d).
\end{equation}

The transform-based variants add only the cost of the corresponding
fixed transform:

\begin{equation}
C_{\mathrm{DCT}}
=
O\left(\frac{Nd^2}{H}\right),
\qquad
C_{\mathrm{WHT}}
=
O(Nd\log d_k).
\end{equation}

For the Shearlet-inspired Fourier filterbank,

\begin{equation}
C_{\mathcal{S}}
=
O\left(dN_p\log N_p + HMN_p\right),
\end{equation}

where $N_p=N-1$ is the number of image patches and $M$ is the number
of filterbank elements.

Thus, the proposed units reduce the projection cost while retaining
the $O(N^2d)$ token-interaction complexity of standard attention.

\begin{table}[ht]
\centering
\caption{Computational complexity of the attention units.}
\label{tab:complexity}
\begin{tabular}{lc}
\toprule
\textbf{Unit} & \textbf{Complexity} \\
\midrule
QKV & $O(4Nd^2+2N^2d)$ \\
QKK & $O(3Nd^2+2N^2d)$ \\
QK-DCTK & $O(3Nd^2+2N^2d+Nd^2/H)$ \\
QK-WHTK & $O(3Nd^2+2N^2d+Nd\log d_k)$ \\
QK-ShearK & $O(3Nd^2+2N^2d+C_{\mathcal{S}})$ \\
QK-MFQK & $O(3Nd^2+2N^2d)$ \\
\bottomrule
\end{tabular}
\end{table}

\section{Experimental Results}

\subsection{Transform-Based Dual-Head Results}

All transform-based experiments were conducted on the CIFAR-10 dataset
using a compact Mini-ViT architecture. The model uses $32\times32$ input
images with $4\times4$ patches, an embedding dimension of 128, four
Transformer blocks, four attention heads, and an MLP ratio of 2.
All models were trained for 30 epochs under the same training settings
to ensure a fair comparison between the attention configurations.

Table~\ref{tab:transform_results} reports the classification accuracy
obtained using the regular attention unit of the Transformer and the
proposed dual-headed variants based on orthogonal transforms and the
filterbank-based Shearlet representation. Results are reported as mean
test accuracy and standard deviation across repeated runs.

Table~\ref{tab:transform_results} reports the classification accuracy
obtained using the regular attention unit of the transformer and various orthogonal transforms, and the filterbank-based Shearlet transform in different
locations of the proposed dual-headed attention unit. Results are
reported as mean test accuracy and standard deviation across repeated
runs.

\begin{table}[ht]
\centering
\caption{Test accuracy and parameter count of the evaluated attention configurations. }
\label{tab:transform_results}

\begin{tabular}{lcc}
\toprule
\textbf{Configuration} &
\textbf{Parameters} &
\textbf{Test Accuracy (\%)} \\
\midrule

Regular Attention: $A(Q,K)V$
    & 546,186
    & $76.035 \pm 0.530$ \\

$A(Q,K)K$
    & 480,138
    & $76.155 \pm 0.035$ \\

\midrule

$A(Q,K)\operatorname{DCT}(K)$
    & 480,138
    & $76.140 \pm 0.806$ \\

$A(Q,K)\operatorname{HT}(K)$
    & 480,138
    & $76.830 \pm 0.764$ \\

\midrule

$A(Q,K)\mathcal{S}(K)$
    & 480,634
    & $\mathbf{77.320 \pm 0.354}$ \\

$A(Q,\mathcal{S}(K))\mathcal{S}(K)$
    & 480,634
    & $77.310 \pm 0.976$ \\

$A(\mathcal{S}(Q),\mathcal{S}(K))K$
    & 480,634
    & $76.970 \pm 1.075$ \\

$A(\mathcal{S}(Q),\mathcal{S}(K))\mathcal{S}(K)$
    & 480,634
    & $77.145 \pm 0.757$ \\

\bottomrule
\end{tabular}
\end{table}

%The top row in Table 2 is the regular transformer. 
Among the tested Shearlet configurations,
$A(Q,K)\mathcal{S}(K)$ achieved the highest mean test accuracy of
$77.320\%$, with a standard deviation of $0.354$. A nearly identical
mean accuracy of $77.310\%$ was obtained when the Shearlet-transformed
key representation was used both in the attention-score computation
and in the output. Transforming both $Q$ and $K$ also produced
competitive results, although the corresponding configurations
exhibited somewhat larger run-to-run variation.

These results suggest that the directional and multiscale information
provided by the filterbank representation can be effectively
incorporated into a two-projection attention unit. In particular,
using the original $Q$ and $K$ to determine the attention weights
while replacing the conventional value representation by
$\mathcal{S}(K)$ provided the highest mean accuracy among the tested
Shearlet configurations.

Walsh-Hadamard transform (WHT) based dual-headed attention unit is the computationally most efficient one because the WHT is binary, i.e., $HT(K)$ can be implemented without performing any multiplications. Furthermore, it achieves better results than the ordinary attention unit based transformer.

\begin{table*}[t]
    \centering
    \caption{Comparison of conventional ViT and normalized MF--tanh
    attention. Top-1 accuracy is the best accuracy on the fixed internal
    validation split from one training run (seed 0). ``Mult.'' counts
    general multiplications and excludes sign, absolute-value, addition,
    tanh, reduction, and division operations. C10 represent the CIFAR10 dataset and Tiny-IN represent Tiny ImageNet dataset.}
    \label{tab:mf_tanh_vit_results}
    \begin{tabular}{llccccc}
        \toprule
        Dataset & Attention & Top-1 (\%) & Top-5 (\%) & Params (M)  & Mult. (M)  \\
        \midrule
        C10
          & Softmax ViT         & \textbf{81.92} & \textbf{98.82}     & 2.694                & 182.85          \\
          & Normalized MF--tanh & 77.80  & 98.70        & \textbf{2.471} & \textbf{168.47}  \\
        \midrule
        Tiny-IN
          & Softmax ViT         & \textbf{40.39} & \textbf{65.35}     & 2.758                & 184.66          \\
          & Normalized MF--tanh & 35.40    & 60.32    & \textbf{2.536}  & \textbf{170.28}  \\
        \bottomrule
    \end{tabular}
\end{table*}

In Table~\ref{tab:mf_tanh_vit_results}, we use the same compact ViT~\cite{dosovitskiy2020image} configuration in all comparisons: six
transformer blocks, embedding dimension $D=192$, three attention heads,
and an MLP hidden dimension of 768. CIFAR-10~\cite{krizhevsky2009learning} uses $4\times4$ patches,
whereas Tiny-ImageNet~\cite{deng2009imagenet} uses $8\times8$ patches; both settings produce 65
tokens including the class token. All models are trained from scratch
for 100 epochs with AdamW, a batch size of 128, an initial learning rate
of $3\times10^{-4}$, weight decay 0.05, five warm-up epochs, and cosine
learning-rate decay. We use fixed 45,000/5,000 and 90,000/10,000
training/selection splits for CIFAR-10 and Tiny-ImageNet, respectively.
The official test sets are not used for model selection.

Across both datasets, normalized MF--tanh attention removes 222,336
parameters and 14.38 million general multiplications from the six-block
model. Its accuracy is lower than that of softmax ViT in the present
single-seed experiments, with gaps of 4.12 and 4.99 percentage points on
CIFAR-10 and Tiny-ImageNet, respectively. These results establish a
consistent efficiency--accuracy tradeoff: the proposed construction
eliminates every learned value projection while retaining most of the
baseline accuracy. Multi-seed evaluation on the official evaluation
sets is required before reporting final mean and standard-deviation
results.

\section{Conclusion}

We introduced a dual-headed Transformer formulation that removes the
independently learned value projection and constructs the output from
the key representation or from fixed transforms of the key. This
reduces the number of learned projection parameters while preserving
the standard token-to-token attention mechanism.

Among the investigated DCT, WHT, QKK, and Shearlet-inspired
configurations, the multiscale directional filterbank produced the
highest mean accuracy. In particular,
$A(Q,K)\mathcal{S}(K)$ achieved $77.32\%$ test accuracy, indicating
that directional and multiscale information can provide a useful
alternative to a learned value projection. The results also show that
the transform can be introduced without modifying the conventional
$QK^T$ attention map. WHT based projection method is the computationally most efficient one.

These findings motivate further investigation of fixed and
filterbank-based operators for reducing the parameter and memory
requirements of Transformer attention. Future work will evaluate the
proposed dual-headed units on larger image datasets and architectures,
and will investigate more efficient filterbank designs and
hardware-oriented implementations.

\newpage

\bibliographystyle{plain}
\bibliography{references}

\end{document}